\documentclass{article}

 \usepackage[preprint]{neurips_2026}

\usepackage[utf8]{inputenc} 
\usepackage[T1]{fontenc}    
\usepackage{hyperref}       
\usepackage{url}            
\usepackage{booktabs}       
\usepackage{amsfonts}       
\usepackage{amsmath}
\usepackage{nicefrac}       
\usepackage{microtype}      
\usepackage{xcolor}         
\usepackage{hyperref}
\hypersetup{
    colorlinks=true,
    linkcolor=blue,
    filecolor=magenta,
    urlcolor=cyan,
}

\usepackage[pdftex]{graphicx}
\usepackage{tabularx}
\usepackage{array}

\usepackage{graphicx}
\usepackage{booktabs}
\usepackage{multirow}
\usepackage[rightcaption]{sidecap}
\sidecaptionvpos{figure}{c}

\usepackage{siunitx}
\DeclareSIUnit\percent{\%}

\title{Toward SLM-based agentic task-tool intent matching}
\workshoptitle{SLM-Agents: 1st Workshop on SLMs for Agentic Systems}

\author{%
  \textbf{Chiara Troiani \quad
  Arash Salarian \quad
  Majed El Helou \quad
  Benjamin Ryder} \\
  \textbf{Jean Diaconu \quad
  Herv\'e Muyal \quad
  Marcelo Yannuzzi} \\
  {\normalfont Cisco Systems, Switzerland} \\
  {\normalfont\texttt{\{chtroian, asalaria, melhelou, beryder,}} \\
  {\normalfont\texttt{jdiaconu, hmuyal, mayannuz\}@cisco.com}}
}

\begin{document}

\maketitle

\begin{abstract}
Tool-equipped AI agents use tool calls to access data and act on external systems. Horizontal growth of agentic systems increases the number of these interactions, and further motivates the need for automated, per-call oversight that can operate at low latency and/or on-prem. Conventional authorization schemes can determine whether an agent is allowed to invoke a tool, but cannot assess the agent’s underlying cognition, specifically, whether the tool selection represents a logical, relevant step toward satisfying the intent of the task or not. Consequently, an allowed call may still deviate from the task's intent: a rogue agent might deviate the calls or nudge other agents to make a combination of calls that would not align with the intent of the task. Therefore, every call needs to be verified. In this study we investigate the applicability of Small Language Models (SLMs) to this purpose: an SLM functions as a task-tool relevance classifier that evaluates every selected tool independently against the assigned task and returns a relevance signal for downstream enforcement. Equipped with a novel dataset with multi-tool tasks whose required tools span distinct Model Context Protocol (MCP) servers, we used prompt-optimization, supervised fine-tuning, and reinforcement learning through GRPO to optimize and specialize SLMs.
\end{abstract}

\section{Introduction}

Agents within a multi-agent system (MAS) rely on tool usage as the main method to interact with the environment. As these systems scale out, a trajectory highlighted by recent discourse on the Internet of Agents~\cite{fleming2025layered,chen2025internet} and its next evolution into the Internet of Cognition~\cite{pandey2026scaling,riordan2026multi}, the research into fending off threats and ensuring the security of these systems is also advancing rapidly. 

Our work extends Task Based Access Control (TBAC) to intent-based TBAC: broadly, TBAC dynamically activates, tracks, suspends, consumes, and revokes permissions as the authorization steps of a task progress~\cite{thomas1998tbac}. In other words, a permission is available only when required by the current stage of an active task, and only for its authorized lifetime and usage. However, the original TBAC does not determine whether a tool semantically matches a user’s or an agent’s natural language intent. Furthermore, classical TBAC assumes that the tasks, workflow steps, dependencies, and permissions have been explicitly modeled. In agentic systems, however, due to the enormous flexibility of natural language, explicit modeling of these components ahead of time becomes impossible.

In order to assess semantic relevance of a tool to a given task a system needs to inspect both the subject query as well as the tool descriptions. Cloud-based, frontier LLMs with reasoning capability are effective for analyzing these calls (see the results section). However, in many organizations, concerns over privacy, cost, latency or organizational policies limit the practical applications of these models. The hypothesis for our study is that optimizing an SLM for this purpose is sufficient to achieve high accuracy while satisfying the operational constraints. \let\thefootnote\relax\footnote{Code and data are available at \url{https://github.com/outshift-open/outshift-casa-slm}}

\section{Method}

\subsection{Methodology Foundation}
Let \(\mathcal{S}\) denote the set of MCP servers parsed and prepared for our problem's underlying data. For \(s\in\mathcal{S}\), let \(\mathcal{T}_s\) be its entire tool catalog, with \(t\in\mathcal{T}_s\) denoting one tool (the tool's name and its metadata). We represented tool candidates as server-tool pairs \(c=(s,t)\), keeping identically named tools on different servers distinct. We created tasks that should require their intended sets of tools for execution. For a task \(q\), let \(G(q)\) be its annotated grounding set of candidate tools, \(R(q)=\{s\in\mathcal{S}:\exists\,t\in\mathcal{T}_s,\,(s,t)\in G(q)\}\) the represented servers, and \(N_q=|G(q)|\) the number of tools paired with $q$ that are relevant to its execution. We set the task-tool relevance matching label \(y(q,c)=1\) if \(c\in G(q)\) and \(y(q,c)=0\) otherwise. Our generated data thus satisfy:
\[
N_q=|G(q)|=|R(q)|, \qquad N_q\in\{2,3\}.
\]
Hence, each task originally has two or three grounding tools, exactly one from each represented server, and is cross-MCP with \(|R(q)|\geq2\). In contrast, the prior work within ASTRA drew all grounding tools from one unique server $s$~\cite{elhelou2025delegated, elhelou2026hybrid}, which makes for lower-complexity tasks. We retained its data samples and pairwise targets, as well as similar data processed and derived from TOUCAN~\cite{xu2025toucan}, while additionally extending with our novel tasks that have tools spanning across multiple servers. We simulate intent-breaking matches with \(y(q,c)=0\) by associating $q$ to candidate tools $c\notin G(q)$, while preserving the total number of distinct tool servers $N_q$. We assigned \(y(q,c)\) labels separately for each candidate $c$ of a task $q$ to enable fine-grained data curation, and a group identifier links all the different candidates from the same multi-tool task.

\textbf{Classification objective.}
Given our labeled task-candidate pairs
$\mathcal{D} = \{(q_i, c_i, y_i)\}$,
we formulate task-tool relevance as a binary classification problem.
The classifier receives the task $q$ and the
candidate tool's name and description and produces a structured output
from which a boolean prediction
$\hat{y}(q,c) \in \{0,1\}$ is parsed.
A false positive allows an irrelevant tool, whereas a false negative
rejects a tool required by the task; broken outputs are logged
as parsing failures.

\textbf{Operational objective.}
Due to the needs of the downstream security task, we set the often used \num{95}\% goal~\cite{al2022impact,sallam2026intrusion} for the accuracy and $F_1$ score.

\subsection{Dataset Setup}
We used a server catalog for data generation containing \num{352} tools from \num{12} distinct MCP servers. We divided it into two server-disjoint pools: \num{194} tools from \num{8} servers and \num{158} tools from \num{4} servers, to serve as non-overlapping validation or test sets. Grounding and alternative tools always remain within their assigned pool, with no task $q$ causing any overlap.

For each target size \(N\in\{2,3\}\), the first stage uses every tool in the pool equally often as an anchor. Given an anchor, it samples \(N-1\) distinct servers uniformly at random without replacement from the other servers in the pool, then samples one tool uniformly at random from each chosen server. At the sampling front, anchor coverage is exact; eligible server combinations are equally likely, as are tools within a chosen server. We note that after sampling the realized frequencies can change during generation and curation, and probabilities across all tools differ because server catalogs have different sizes $|\mathcal{T}_s|$. The sampled tool sets are then processed with a similar pipeline as ASTRA's~\cite{elhelou2025delegated}, using GPT-5.2 to produce natural-language tasks $q$ matching with the selected tools' names and descriptions, without exposing the tool-side input schema to avoid over-fitting.

The second stage leads to three types of \(N\)-tool candidate sets for sampled tasks. A \emph{correct} set equals \(G(q)\). A \emph{wrong} set replaces grounding tools with different tools from the same server. A \emph{null} set draws candidates from a server outside \(R(q)\) but within the same pool, such that its servers and the grounding servers do not overlap. We enforced the \emph{wrong} and \emph{null} logic in our cross-server setting, notably by masking all $G(q)$ servers from the sampling pool of \emph{null} samples. Ultimately, both \emph{wrong} and \emph{null} are mapped to an \emph{irrelevant} task-tool label \(y(q,c)=0\). The generation and sampling stages do not mix tool types in a given set, however, curation may introduce heterogeneous candidate sets for a given task, which is why we preserved candidate-level labels. Each candidate set is represented as \(N\) task-tool rows for relevance curation. For consistency, we performed a similar curation process as the one proposed by ASTRA~\cite{elhelou2025delegated}. We retained a row only when all recorded curators for that row agreed, and kept only groups containing all expected \(N\) rows. Out of our \num{9289} curated rows, \num{7811} formed \num{3155} complete groups, and \num{1478} rows in \num{996} incomplete groups were removed.

\subsection{Optimization Approach}

We applied the same sequential optimization strategy to two models of Gemma~3 family (1B and 4B):
\[
\text{Base} \rightarrow \text{GEPA} \rightarrow \text{SFT}
\rightarrow \text{GRPO}.
\]
GEPA~\cite{agrawal2026gepa} produced the prompt used for SFT, and
GRPO~\cite{shao2024deepseekmath} was initialized from the selected SFT
checkpoint. Prompt and checkpoint selection used validation $F_1$ for the
relevant class. The test set was evaluated only after all experimental choices
had been fixed and shared no MCP servers, tools, or generated tasks with the
training or validation splits. Each example contained a task, a candidate-tool
name and description, and a target response of the form
\texttt{\{"reasoning": "...", "appropriate": true/false\}}.

\textbf{GEPA.}
To optimize agent's prompt, we used GEPA~\cite{agrawal2026gepa}. We selected a random, balanced subset of the training and validation data (\num{500} and \num{250} samples, respectively). GPT-5.6 Sol was the optimizer/reflection LLM, with a reflection mini-batch size of \num{35}. For reporting the test results, however, we used the entire test dataset.

\textbf{SFT.}
Using the optimized prompt, we trained LoRA~\cite{hu2021lora} adapters
($r=32$, $\alpha=64$) on \num{4404} class-balanced, single-tool ASTRA examples
and selected the checkpoint on \num{1136} validation examples. We treated single-tool tasks as a lower-complexity curriculum stage because they expressed direct task-tool relationships without the compositional dependencies of multi-tool tasks. We masked prompt tokens and minimized cross-entropy over the model response. Training used AdamW with a peak learning rate of \num{1.9e-4}, \num{10}\% warmup, three epochs, an effective batch size of \num{64}, and a maximum sequence length of \num{1024} in \texttt{bf16}. To select the hyper-parameters, we used TPE sampler (Tree-structured Parzen Estimator) in Optuna~\cite{optuna_2019}, optimizing for validation set's $F_1$.

\textbf{GRPO.}
Starting from the $F_1$-selected SFT checkpoint, we applied GRPO to a class-balanced
mixture of single-tool, same-MCP multi-tool, and cross-MCP multi-tool examples
(\num{9485} training and \num{2927} validation). The deterministic reward combined
schema compliance, label correctness, and a penalty for excessive reasoning;
false positives and false negatives were penalized equally, and invalid outputs
received an additional penalty. We did not use an LLM-based reasoning judge.
We updated the LoRA adapters and sampled \num{16} responses per prompt at
temperature \num{1.1}. We optimized the Dr.\ GRPO~\cite{liu2025drgrpo} objective with
sequence-level importance sampling, batch-scaled group-relative advantages,
and a KL penalty of $\beta=10^{-3}$. Training used a learning rate of
\num{e-5} for three epochs at an effective batch size of \num{256} in
\texttt{bf16}; prompts and completions were capped at \num{1024} and \num{192} tokens,
respectively.

\section{Experimental Setup}
We evaluated the Gemma~3 1B and 4B model sizes and sought the
smallest evaluated model that can reach this target through progressive
specialization. The training data were balanced with respect to the binary relevance label,
whereas the validation and test sets retained their curated class
distributions. Prompt, hyperparameter, and checkpoint selection used only the
validation split, with $F_1$ for the relevant class as the primary selection
metric. The held-out test set was evaluated only after all experimental choices
had been fixed. Aggregate performance is reported in Table~\ref{tab:results-table}.

Inference was performed with vLLM using a fixed decoding configuration across
all four conditions. We report parse success, end-to-end (E2E) accuracy, accuracy,
precision, recall, $F_1$, false-positive rate, and false-negative rate.
E2E accuracy was computed over the complete test set, with malformed or
unparseable outputs counted as incorrect predictions. All other classification
metrics were computed from successfully parsed Boolean predictions, with failed parses reported separately. We obtained \num{95}\% bias-corrected and
accelerated confidence intervals from \num{10000} paired bootstrap resamples. A server with a single H100 GPU was used for all three steps of the optimization.

\section{Results and Discussion}
\label{sec:results}

\textbf{Frontier LLMs.}
GPT-5.6~Sol achieved \num{98.58}\% accuracy and \num{98.86}\% $F_1$ on the
held-out test set, with false-positive and false-negative rates of \num{1.42}\% and
\num{1.41}\%, respectively, and no parse failures; exceeding our operational threshold of 95\% accuracy. 

\begin{table*}[t]
\caption{Task-tool relevance classification and end-to-end performance on
the held-out test set for Gemma~3~4B and Gemma~3~1B. Scores are percentages;
brackets give \num{95}\% confidence intervals. Bold indicates
the best result for each metric.}
\vspace{2mm}
\label{tab:results-table}
\centering
\footnotesize
\renewcommand{\arraystretch}{0.8}

\begin{tabular*}{\textwidth}{@{\extracolsep{\fill}}clcccc@{}}
\toprule
Model & Metric & Base & GEPA & SFT & GRPO \\
\midrule

\multirow{8}{*}{%
  \rotatebox[origin=c]{90}{\textbf{Gemma~3~1B}}%
}
  & Failed parses $\downarrow$
  & 103
  & \textemdash
  & 175
  & 1 \\

  & E2E Acc. $\uparrow$
  & 60.94 [59.97, 61.91]
  & \textemdash
  & 78.63 [77.80, 79.42]
  & 89.60 [88.99, 90.18] \\

  & Accuracy $\uparrow$
  & 61.59 [60.64, 62.59]
  & \textemdash
  & 80.06 [79.26, 80.87]
  & 89.60 [88.99, 90.20] \\

  & Precision $\uparrow$
  & 72.53 [71.30, 73.69]
  & \textemdash
  & 94.54 [93.83, 95.16]
  & 94.61 [94.01, 95.19] \\

  & Recall $\uparrow$
  & 61.80 [60.56, 63.04]
  & \textemdash
  & 72.13 [70.99, 73.25]
  & 88.39 [87.55, 89.16] \\

  & $F_1$ $\uparrow$
  & 66.74 [65.72, 67.75]
  & \textemdash
  & 81.82 [81.01, 82.63]
  & 91.39 [90.85, 91.91] \\

  & FPR $\downarrow$
  & 38.76 [37.21, 40.36]
  & \textemdash
  & 6.86 [6.07, 7.74]
  & 8.37 [7.50, 9.30] \\

  & FNR $\downarrow$
  & 38.20 [36.96, 39.44]
  & \textemdash
  & 27.87 [26.75, 29.01]
  & 11.61 [10.84, 12.45] \\

\midrule

\multirow{8}{*}{%
  \rotatebox[origin=c]{90}{\textbf{Gemma~3~4B}}%
}
  & Failed parses $\downarrow$
  & \textbf{0}
  & 9
  & 1
  & \textbf{0} \\

  & E2E Acc. $\uparrow$
  & 87.48 [86.81, 88.13]
  & 91.58 [91.01, 92.11]
  & 93.78 [93.29, 94.25]
  & \textbf{96.13} [95.73, 96.50] \\

  & Accuracy $\uparrow$
  & 87.48 [86.81, 88.13]
  & 91.66 [91.11, 92.21]
  & 93.79 [93.29, 94.25]
  & \textbf{96.13} [95.73, 96.50] \\

  & Precision $\uparrow$
  & 95.29 [94.69, 95.82]
  & 92.17 [91.46, 92.80]
  & 95.75 [95.23, 96.24]
  & \textbf{96.75} [96.30, 97.18] \\

  & Recall $\uparrow$
  & 84.11 [83.16, 85.01]
  & 94.70 [94.12, 95.24]
  & 94.23 [93.62, 94.77]
  & \textbf{97.06} [96.61, 97.46] \\

  & $F_1$ $\uparrow$
  & 89.35 [88.73, 89.93]
  & 93.42 [92.96, 93.86]
  & 94.98 [94.57, 95.36]
  & \textbf{96.90} [96.58, 97.20] \\

  & FPR $\downarrow$
  & 6.92 [6.14, 7.77]
  & 13.39 [12.34, 14.55]
  & 6.95 [6.17, 7.81]
  & \textbf{5.42} [4.71, 6.17] \\

  & FNR $\downarrow$
  & 15.89 [14.99, 16.84]
  & 5.30 [4.76, 5.88]
  & 5.77 [5.23, 6.38]
  & \textbf{2.94} [2.54, 3.39] \\

\bottomrule
\end{tabular*}
\end{table*}

\begin{SCfigure}[2][tbp]

  \centering
  \includegraphics[
    width=0.35\linewidth, trim={10 15 10 20}, clip
  ]{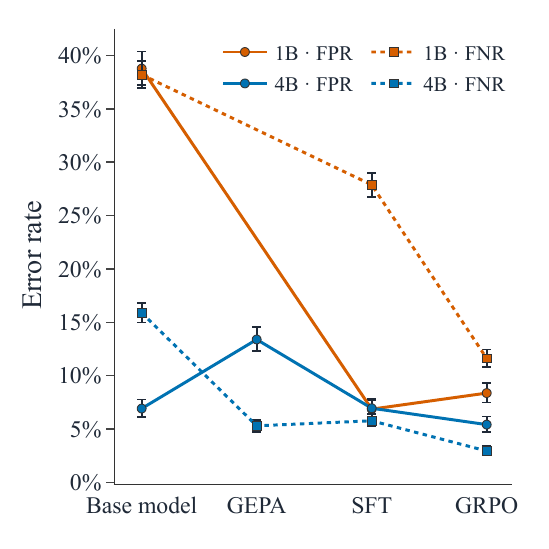}

  \caption{False-positive rate (FPR) and false-negative rate (FNR) on the
    held-out test set (\num{9736} samples) for Gemma~3~1B and Gemma~3~4B across successive
    stages of the specialization pipeline: Base, GEPA, SFT, and GRPO.
    GRPO achieves the lowest rates for Gemma~3~4B with FPR of \num{5.42}\% and FNR of
    \num{2.94}\%. Because the stages were applied sequentially, the results
    represent cumulative improvements rather than independent ablations. For Gemma~3 1B, an em dash indicates that GEPA produced no validation improvement and therefore no separate configuration was selected for test evaluation.
  }
  \label{fig:fpr-fnr-training-data}
\end{SCfigure}

\textbf{SLM: Overall performance.}
Specialization improved both model sizes.
For Gemma~3 1B, GEPA did not improve validation performance, so the seed prompt was retained and no separate GEPA
result was evaluated on the held-out test set. SFT increased accuracy from
\num{61.59} to \num{80.06} and $F_1$ from \num{66.74} to \num{81.82}, although parse failures
increased from \num{103} to \num{175}, limiting E2E accuracy to \num{78.63}. GRPO
largely resolved this issue, reducing parse failures to one while increasing
E2E accuracy to \num{89.60} and $F_1$ to \num{91.39}.
For Gemma~3 4B, performance improved at every stage of the sequential pipeline
(Table~\ref{tab:results-table}). GEPA increased E2E accuracy and $F_1$ by \num{4.10} and \num{4.07} percentage points (pp), respectively; SFT added \num{2.20} and \num{1.56} pp,
and GRPO a further \num{2.35} and \num{1.92} pp, yielding cumulative gains of \num{8.65} and \num{7.55} pp.

\textbf{Error trade-off across stages.}
The stages affected false positives and false negatives differently
(Figure~\ref{fig:fpr-fnr-training-data}). 
For Gemma~3~1B, SFT substantially reduced the false-positive rate from \num{38.76}
to \num{6.86}, while GRPO reduced the false-negative rate from \num{27.87} to \num{11.61},
at the cost of a modest increase in false positives to \num{8.37}.
For Gemma~3~4B, GEPA reduced the false-negative rate from \num{15.89} to \num{5.30}, but increased the false-positive rate from \num{6.92} to \num{13.39}. SFT restored selectivity, reducing false positives to \num{6.95} while
retaining most of the recall gain. GRPO produced the best balance, with
false-positive and false-negative rates of \num{5.42} and \num{2.94}, respectively.
Thus, GEPA prompt optimization, SFT, and GRPO made complementary contributions.

\section{Conclusion}
While frontier LLMs can meet the \num{95}\% operational target (GPT-5.6~Sol achieved an accuracy of \num{98.58}\%), both SLMs initially performed well below this threshold, with E2E accuracies of \num{60.94}\% and \num{87.48}\% for Gemma~3 1B and 4B respectively. 
Sequential specialization, however, substantially improved the performance of both model sizes.  
After GRPO, Gemma~3~1B reached \num{89.60}\% E2E accuracy, remaining below the operational target, whereas Gemma~3~4B reached \num{96.13}\%, slightly surpassing it. Although our evaluation is limited to synthetic, LLM-curated task-tool pairs and two model sizes from a single family, these results support our hypothesis that specialized SLMs can provide a viable operational alternative to frontier models for intent-based TBAC in agentic systems.

\newpage
\bibliographystyle{unsrt}
\bibliography{references}

\end{document}